\documentclass[10pt,twocolumn,letterpaper]{article}

\usepackage{cvpr}              
\usepackage{multirow}
\usepackage{listings}
\usepackage{xcolor}
\usepackage{makecell}

\definecolor{cvprblue}{rgb}{0.21,0.49,0.74}
\usepackage[pagebackref,breaklinks,colorlinks,allcolors=cvprblue]{hyperref}

\title{Diffusion Restoration Adapter for Real-World Image Restoration}

\author{Hanbang Liang, Zhen Wang, Weihui Deng\\
{\tt\small \{lianghanbang, wangzhen3560, weihuideng\}@bytedance.com}
}

\begin{document}

\twocolumn[{%
\renewcommand\twocolumn[1][]{#1}%
\maketitle
\centering
\includegraphics[width=1\textwidth]{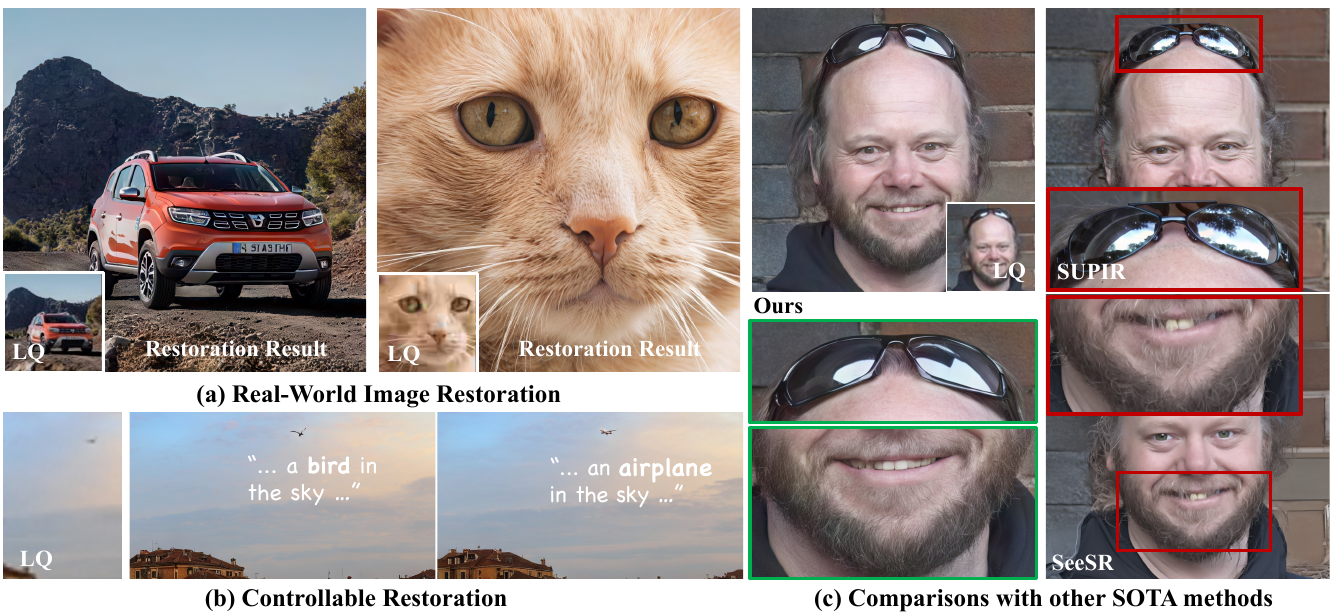}
\captionof{figure}{The proposed Diffusion Restoration Adapter delivers high-quality image restoration. Our proposed methods can utilize descriptive prompts to achieve controllable restoration, as shown in (b). Our methods show competitive on qualitative comparisons with other SOTA methods.  \vspace{1em}}
\label{fig:teaser}
}]

\begin{abstract}
Diffusion models have demonstrated their powerful image generation capabilities, effectively fitting highly complex image distributions. These models can serve as strong priors for image restoration. Existing methods often utilize techniques like ControlNet to sample high quality images with low quality images from these priors. However, ControlNet typically involves copying a large part of the original network, resulting in a significantly large number of parameters as the prior scales up. In this paper, we propose a relatively lightweight Adapter that leverages the powerful generative capabilities of pretrained priors to achieve photo-realistic image restoration. The Adapters can be adapt to both denoising UNet and DiT, and performs excellent. 
\end{abstract}    
\section{Introduction}
Image restoration aims to recover high-quality images from degraded observations.  From early on, most works have focused on establishing a mapping relationship from low-quality (LQ) image to high-quality (HQ) images using neural networks, including CNNs and Transformer-based architectures \cite{zhang2017learning,zhang2021plug,liang2021swinir,wang2021real,zhang2021blind,Zamir_2022_CVPR,Zhou_2024_CVPR}. These methods often incorporated GANs \cite{goodfellow2014generative} as auxiliary losses to improve the quality of the generated images. 

Recently, the concept of generative prior for image restoration has gained traction \cite{wang2023exploiting,lin2023diffbir,menon2020pulse,yang2021gan,wang2021towards,zhao2022rethinking,zhou2022towards,yu2024scaling}. This approach leverages the powerful generative capabilities of pretrained priors, using low-quality images as conditions to generate high-quality images. Previous works have utilized GANs \cite{menon2020pulse,yang2021gan,wang2021towards}, such as StyleGAN \cite{karras2019style,karras2020analyzing}.
Many StyleGAN-based works operate within the latent space of StyleGAN, achieving controllable generation by modifying the latent code\cite{karras2020analyzing, shen2020interfacegan, abdal2021styleflow,liang2021ssflow,tewari2020stylerig,harkonen2020ganspace}. This process often involves methods related to GAN inversion\cite{gu2020image,menon2020pulse,abdal2019image2stylegan,pan2021exploiting,bau2020semantic}. Despite the inefficiency of GAN inversion, these methods have shown promising results in face image restoration. However, due to the limitations of GANs, these methods perform well on aligned face datasets but struggle to fit distributions effectively on large-scale and diverse datasets.

The rise of diffusion models has opened up new possibilities for fitting distributions on large-scale and diverse datasets \cite{ho2020denoising, podell2023sdxl, chen2023pixart, peebles2023scalable, esser2024scaling}. By combining Latent Diffusion Models (LDM) \cite{rombach2022high} and using text as a condition, the ability of models to generate high-quality images has been further enhanced. As a result, diffusion models have become the most widely used pretrained prior for image restoration today.

Using pretrained diffusion models as generative priors for image restoration primarily involves fine-tuning a module that enables the prior to perform conditional generation, using the low-quality (LQ) image as a condition. ControlNet \cite{zhang2023adding}, a technique for controlling diffusion generation, has achieved excellent results in conditional generation tasks, using conditions such as edge, depth map, and binary mask to control the generation effectively. Naturally, there has been exploration into using ControlNet for image restoration as well. However, ControlNet typically has a large number of parameters, which imposes significant training and inference burdens, especially as the model scales up.

In contrast to conditions like depth maps and edges, LQ images usually contain more meaningful features. The process of generating high-quality (HQ) images using the prior is more similar to finding a sample in the distribution that closely matches the LQ image. We believe there can be a more simple and effective way to achieve restoration using a prior. Therefore, we propose that using a smaller module instead of ControlNet can be more effective for image restoration with priors.

To address this, this paper introduces a Restoration Adapter, which is more lightweight than ControlNet and is inserted into the original model. The adapter embeds the LQ image as a condition into the features of the original model to achieve conditional sampling. On the other hand, pretrained priors are typically trained on super large-scale datasets \cite{schuhmann2022laion}. When tuning these priors for image restoration tasks, we usually collect only a small portion of high-quality data for training, and this data may not have appeared in the original dataset. Therefore, we decide to fine-tune the prior simultaneously. Besides, we believe fine-tuning the prior can alleviate our Restoration Adapter's burden, leading to better results. For efficient fine-tuning, we incorporate adapters into the parameters of the pretrained network. 
As DiTs \cite{peebles2023scalable} become more popular compared to denoising UNet, our methods are designed to fit both architectures, as we propose a flexible framework. Moreover, to enhance the balance between fidelity and diversity, we introduce a sampling strategy, which can be easily adapted to various samplers.
As shown in Figure \ref{fig:teaser}, our method achieves high-quality restoration results across various types of images and allows for controllable restoration based on specific prompts. 

In summary, we make following contributions: \textbf{(1).} We propose the Diffusion Restoration Adapter, a framework that integrates an adapter into diffusion denoising networks to achieve image restoration.
\textbf{(2).}  We design specific Restoration Adapter modules for both UNet-based diffusion models and DiTs, to make it more convenient for various diffusion priors.
\textbf{(3).}  A straightforward sampling strategy is introduced to ensure fidelity in restoration for various samplers.
\section{Related works}
\begin{figure*}[t]
  \centering
   \includegraphics[width=0.98\linewidth]{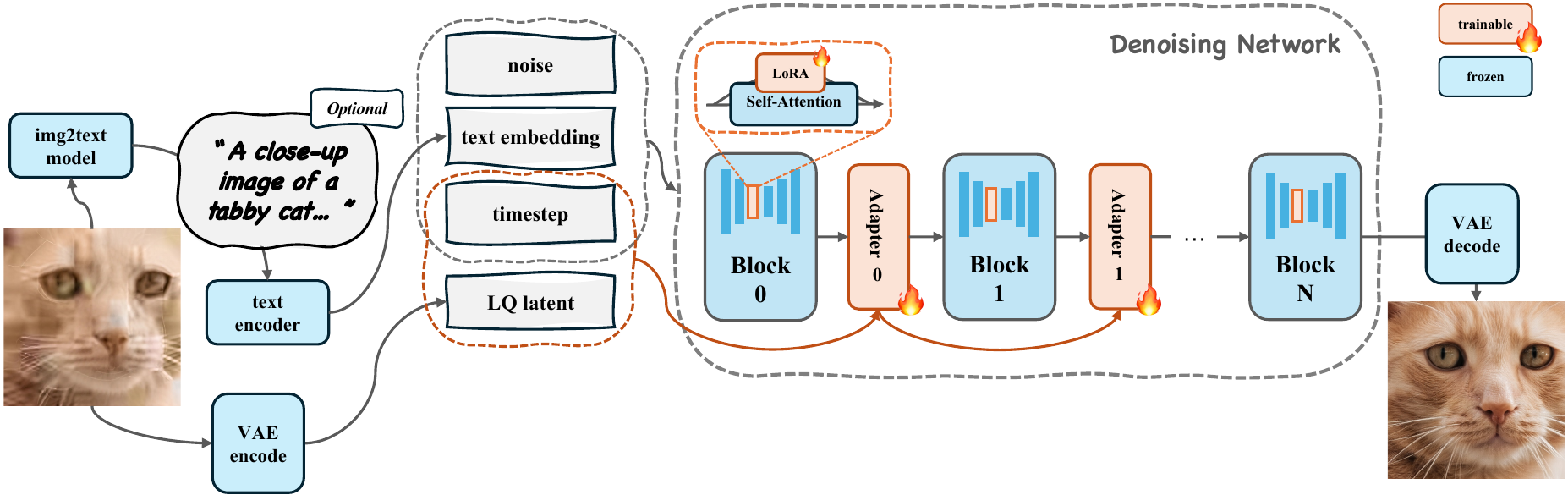}
   \caption{The architecture of the Diffusion Restoration Adapter is integrated within the denoising network. Restoration Adapters are inserted into the network blocks of the original structure. Additionally, Diffusion Adapters are applied to specific parameters, particularly integrating LoRA into the self-attention module within each block.}
   \label{fig:main_arch}
\end{figure*}
\begin{figure}[t]
  \centering
    \includegraphics[width=0.95\linewidth]{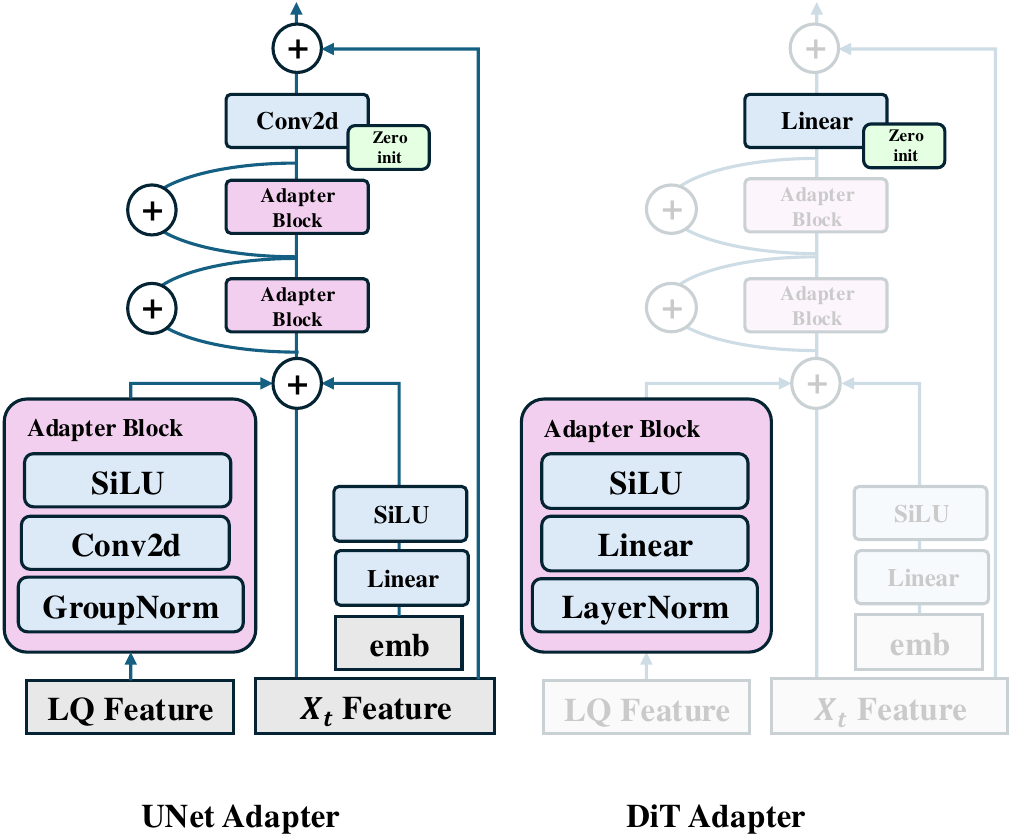}
   \caption{The Restoration Adapter has two variants, specifically designed for the denoising UNet and DiTs. Linear denotes fully-connected layer.}
\label{fig:adapter_arch}
\end{figure}
\subsection{Utilizing Diffusion for Image Restoration}
Diffusion models have made remarkable progress in generating high-quality images. Recently, numerous works \cite{podell2023sdxl, peebles2023scalable, esser2024scaling, chen2023pixart,li2024hunyuan} have proposed methods to improve the quality of generation, especially in text-to-image generation. With their powerful ability to generate high-quality images, researchers have begun to utilize pretrained diffusion models as generative priors to achieve image restoration \cite{yu2024scaling, kawar2022denoising, wang2023exploiting, lin2023diffbir, yang2023pixel, wu2024seesr}. Previous works have utilized other generative model such as Generative Adversarial Networks (GAN) \cite{goodfellow2014generative} as priors \cite{yang2021gan,wang2021towards, menon2020pulse}. Unlike GANs, diffusion models do not generate images in a single forward step. Instead, the generation process combines multiple steps of denoising, making it more challenging to control the generation.

The most widely used neural network architectures in diffusion models are UNet \cite{ho2020denoising} and DiTs (Diffusion Transformers) \cite{peebles2023scalable}. These architectures are designed to effectively capture and process the intricate details during the denoising steps, ensuring high-quality image generation.

Latent Diffusion Models (LDM) \cite{rombach2022high} are another significant advancement in this field. LDMs operate in a compressed latent space rather than the high-dimensional image space, which significantly reduces computational requirements while maintaining high generation quality. This compressed latent space is derived from high-dimensional images using a Variational Autoencoder (VAE) \cite{kingma2013auto}. By leveraging the latent space, LDMs can efficiently model complex image distributions and provide robust priors for various image restoration tasks. Notable examples of LDMs includes StableDiffusion XL \cite{podell2023sdxl}, which is based on the UNet architecture, and StableDiffusion 3 \cite{esser2024scaling}, which utilizes the DiTs \cite{peebles2023scalable} architecture. These models demonstrate the versatility and effectiveness of LDMs in generating high-quality images. DiffBIR\cite{lin2023diffbir} and SUPIR\cite{yu2024scaling} utilize ControlNet to control StableDiffusion, while StableSR \cite{wang2023exploiting} trains a Time-aware Encoder to inject conditions and employs a feature wrapper to enhance the flexibility of generation.

\subsection{Low-Rank adaptation (LoRA)}
Low-Rank Adaptation (LoRA) \cite{hu2021lora} is a technique used to fine-tune large pre-trained models efficiently. By introducing low-rank matrices into the model's architecture, LoRA reduces the number of parameters that need to be adjusted during the fine-tuning process. This approach not only decreases computational costs but also helps in maintaining the model's performance. Such adaptation is widely employed in fine-tuning a pretrained diffusion model with a small set of training data.
\section{Method}
\subsection{Preliminaries}
\paragraph{Diffusion Models.} Diffusion Models are a class of generative models that iteratively transform a simple distribution into a complex one through a series of small, reversible steps. The core idea is to define a forward diffusion process that gradually adds Gaussian noise to the data and a reverse process that denoises it. 

The forward diffusion process can be described by a Stochastic Differential Equation (SDE) \cite{ho2020denoising, song2019generative, song2020score} which outlines how noise is added over time. 
The reverse process aims to recover the original data by reversing the noise addition, which can be modeled using either a Stochastic Differential Equation (SDE) or an Ordinary Differential Equation (ODE). The choice between SDE and ODE depends on the specific methodology and solver (samplers) being used to generate new samples effectively.

\paragraph{Restoration Prior.} Restoration priors are pretrained generative models. In image restoration tasks, low-quality images serve as conditional inputs. By performing conditional sampling within the prior, these models generate corresponding high-quality images. Latent Diffusion Models (LDM) are commonly used priors in image restoration tasks due to their ability to generate high-quality images from text. The StableDiffusion series models are Latent Diffusion Models (LDM) trained on extensive datasets. In this paper, we select StableDiffusion XL and StableDiffusion 3 as priors. The denoising network of StableDiffusion XL is a UNet, whereas StableDiffusion 3 utilizes an MM-DiT, which is similar to DiT. The output layer uses zero initialization to ensures the stable training at the very beginning.

\begin{figure*}[t]
    \centering
    \includegraphics[width=1.0\linewidth]{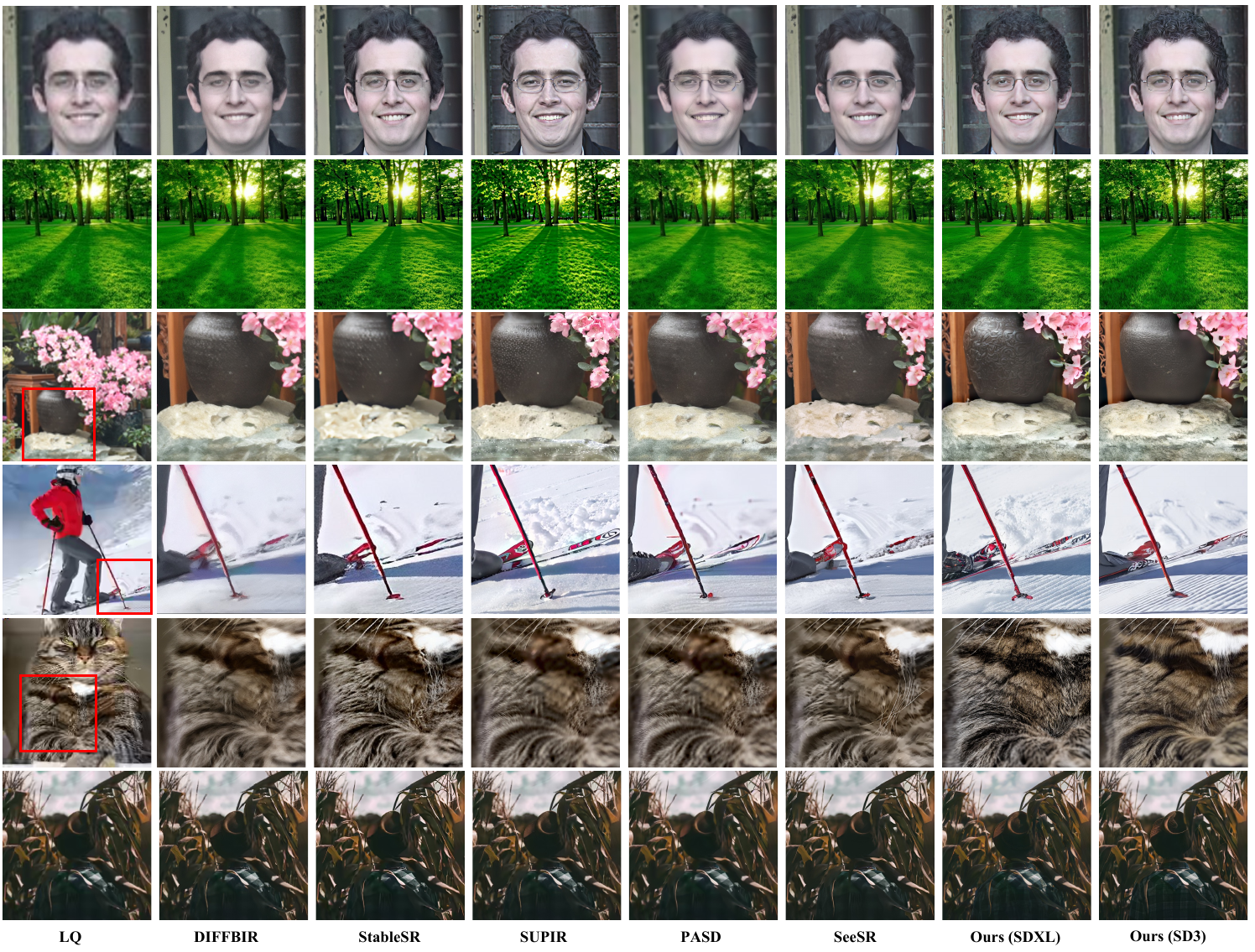}
    \caption{Qualitative comparisons on RealPhoto60 and center-cropped DIV2K validation set. Our methods can perform good results under different kinds of degradations. (Zoom in for details)}
    \label{fig:main_cmp}
\end{figure*}

\subsection{Diffusion Restoration Adapter}
Our proposed Diffusion Restoration Adapter consists of two main components: the Restoration Adapters, which are integrated into the original denoising network architecture, and the Diffusion Adapters, designed to fine-tune specific parameters within the denoising network. We choose pretrained latent diffusion models, specifically StableDiffusion XL (SDXL), which is UNet-based, and StableDiffusion 3 (SD3), which is based on DiTs, as our diffusion priors due to their outstanding performance in text-to-image tasks. For SDXL, there are two stages of generation, we only utilize the base model from the first stage.

\paragraph{Restoration Adapter.}To control the prior generate HQ image from LQ condition input, extra LQ condition processing module is needed. The common choice ControlNet make copies of parts of the original networks and process the inputs in parallel with the main network, then sum the results together at the end. We believe that the extra network processing the additional condition should be more integrated with the main network, rather than operating independently. To address this, we introduce the Restoration Adapter for both UNet-based diffusion and DiTs. Our architecture is shown in Figure \ref{fig:main_arch}. Our proposed approach involves inserting a few adapters into the original denoising network. These adapters are placed after the blocks into which they are being inserted. Note that adapters can be selectively assigned to certain blocks, and not all blocks need to be equipped with an adapter.

In SDXL, there are three downsampling blocks, one mid block, and three upsampling blocks. Sequentially, the ``Block n" (0 $\leq$ n $\leq$ N) shown in Figure \ref{fig:main_arch} corresponds to each of these blocks. For example, ``Block 0" refers to the first downsampling block, ``Block 1" refers to the second downsampling block and so on. In the case of SD3, the denoising network consists of multiple transformer blocks, specifically MM-DiT blocks. Thus, ``Block n" represents the n-th MM-DiT block. After inserted with adapters, the output of each blocks are passed to the corresponding Adapter instead of the next block.

As illustrated in Figure \ref{fig:adapter_arch}, the adapter has two variants tailored for UNet in SDXL and DiTs in SD3. The adapter takes the low-quality (LQ) feature, the denoising feature $X_t$ at timestep $t$, and the timestep embedding $emb$ as inputs. Taking the architecture for UNet as an example, the LQ feature is derived from the previous Adapter's output, with the first Adapter's input being the encoded LQ latent. The LQ feature passes through an Adapter Block, while the time embedding undergoes a linear transformation combined with a SiLU activation to ensure it matches the shape of $X_t$. These features are then summed together. After summation, the feature is processed through two additional Adapter Blocks with residual connections, and finally, it passes through a zero-initialization \cite{zhang2023adding} linear layer and sums with the $X_t$. The output of the Adapter is passed to the next denoising network block. The architecture for DiTs shares a similar design but with a few changes as shown on the right side of Figure \ref{fig:adapter_arch}.

\paragraph{Diffusion Adapter.} Restoration priors generally fit the distribution of extremely large-scale datasets. We fine-tune the prior using a small subset of the training data along with some unseen data. Besides, it's important to fine-tune the original denoising network to alleviate the pressure on the Restoration Adapter. On this basis, we also aim to keep overall efficiency when fine-tuning the prior.
Therefore, to fully leverage the prior's powerful ability to generate high-quality images, thereby enhancing the restoration task, we add adapters to the self-attention module within the denoising network and train them simultaneously with our Restoration Adapter. To keep the approach simple and effective, we have selected the common choice, LoRA \cite{hu2021lora}, as the adapter.

\subsection{Training and Sampling} 
\paragraph{Training Data and Objective.} Training the Diffusion Restoration Adapter requires high-quality images along with their corresponding degraded images. The degraded images are produced by a degradation pipeline following the design from RealESRGAN \cite{wang2021real}. 
Additionally, since the priors are text-to-image generation models, we employ a multi-modal language model to extract short captions for the training images. To enhance robustness across all degradation levels and improve performance with null text input, the text may be set to a null prompt. Additionally, the LQ image can serve as the model's training target, while the input prompts are set to degradation descriptors such as ``low quality," ``low resolution" and specific applied degradations like ``compressed." To improve the performance of our method on SDXL with limited training resources, we employ the degradation-robust encoder \cite{yu2024scaling} for SDXL.

The training process involves using the Diffusion Restoration Adapter as the LQ image condition processor. This processor converts the condition into the intermediate feature of the pretrained network. Consequently, we fine-tune the pretrained prior to form a conditional distribution, with the LQ image as the condition. We freeze the parameters of the pretrained network and conduct training only on our inserted Adapters. The training objective strictly follows the original training process of the prior. For SD3, which is trained with conditional flow matching loss \cite {lipmanflow, esser2024scaling}, we adhere to the same loss function. For SDXL, the training objective follows the loss used in DDPM \cite {ho2020denoising}.

\paragraph{Restoration Sampling Strategy.} Many methods \cite{karras2022elucidating,zhao2024unipc,lu2022dpm,lu2022dpmpp,songdenoising} have been developed to improve the generation quality of diffusion models, generally aiming to enhance diversity and overall quality. However, these methods are not designed for restoration tasks. Consequently, using such sampling schedulers might result in highly unfaithful samples. we propose a straightforward method that can be easily adapted to any sampling technique without altering the core sampler's code. This method adjusts the denoising direction during the generation sampling loop. Similar to Classifier-free guidance \cite{ho2022classifier} and SUPIR \cite{yu2024scaling}, we use the LQ image as guidance. By calculating the unnormalized direction between the denoised latent $z_t$ at timestep $t$ and the LQ latent, we then adjust the denoised latent towards this direction by a factor. This factor is timestep-dependent and is created from a mapping function $g$ :
\begin{equation}
    \label{sampling_strat}
    \begin{aligned}
        z_t &= z_t' + w \cdot g(t, T) \cdot (c_{lq} - z_t')
    \end{aligned}
\end{equation}
where $c_{lq}$ is the LQ latent, $z_t'$ is the output of the original sampling method at time step $t$, $T$ is the total number of sampling timesteps, $w$ serves as a weighting hyperparameter, and $g$ is a mapping function. 
The choice of $g$ should follow the a rule that gives more weight at larger timesteps and vice versa. This is because we want the denoising step to focus more on fidelity during the early denoising loop while considering high quality and diversity in the later loops. For simplicity, we use an piecewise linear mapping function. The function maps the $t$ to $[0, 1)$, with a threshold hyperparameter $a$ determining the piecewise behavior. 
\begin{equation}
    \label{piecewise}
    \begin{aligned}
        g(t, T; a) &=
        \begin{cases}
            \frac{\frac{t}{T} - a}{1 - a} & \text{if } \frac{t}{T} > a \\
            0 & \text{otherwise}
        \end{cases}
    \end{aligned}
    \quad \text{where } 0 < t < T
\end{equation}
By moving the $z_t'$ towards $c_{lq}$, we achieve results with greater fidelity as $z_t'$ gets closer to $c_{lq}$.

\begin{table}[t]
\centering
\resizebox{0.75\linewidth}{!}{
    \begin{tabular}{c| c c c}
    \hline
            Methods & ClipIQA & MUSIQ & ManIQA \\
            \hline
            StableSR & 0.565 & 60.57 & 0.3411 \\
            DiffBIR & 0.6353 & 66.91 & 0.4132 \\
            PASD & 0.6345 & 65.57 & 0.4336 \\
            SUPIR & \underline{0.7027} & 70.42 & \underline{0.5254} \\
            SeeSR & 0.6699 & 71.56 & 0.4763 \\
            Ours (SDXL) & 0.6996 & \textbf{71.83} & 0.4818 \\
            Ours (SD3) & \textbf{0.7067} & \underline{71.62} &  \textbf{0.5292} \\
    \hline
    \end{tabular}
}
\caption{Quantitative comparisons on the RealPhoto60 dataset. The best results are highlighted in \textbf{bold text}, while the second-best results are \underline{underlined}.}
\label{tab:nr_cmp}
\end{table}

\section{Experiments}
\subsection{Configuration}
\paragraph{Training Data.}We collected 300k high-quality images with text descriptions as our training data, including the DIV2K training set \cite{Agustsson_2017_CVPR_Workshops} . Subsequently, we applied the modified degradation pipeline we introduced to generate pairs of low-quality (LQ) and high-quality (HQ) images.

\paragraph{Training and Testing Details.} We train our models using the SDXL and the SD3, as they are UNet-based and DiT-based priors. We set an adapter for each block except for the last upsampling block in SDXL, and for SD3, we insert an adapter after every four blocks. The rank of LoRA is set to 32. We train our model under 512 $\times$ 512 resolution using the AdamW optimizer\cite{loshchilov2017decoupled} with a learning rate of 0.00001, utilizing 300k high-quality images. Additionally, to explore the impact of data size on performance, we train a model with 30k samples which are randomly sampled from the 300k images, for SDXL.

For testing, the hyperparameters are set as follows: T=100, classifier free guidance is set to 7.5, the Restoration Sampling Strategy weight $w$ is set to 0.05 and the threshold hyperparameter $a$ is set to $0.8$. We select InternVL \cite{chen2024internvl} as the image to text model, to extract short descriptive prompts of LQ images.

\paragraph{Testing Data and Evaluation Metrics.} Following previous works on the evaluation of restoration quality, we synthesize LQ images from high-resolution and HQ mages. For the HQ image data, we select the DIV2K Validation set \cite{Agustsson_2017_CVPR_Workshops} as our synthesis image source. The images are resized and center-cropped into 1024 $\times$ 1024. To test the restoration ability on close-up image, we randomly cropped the image pairs to 512 $\times$ 512 for testing the robustness on close-up images. 
With high-quality data and synthesized low-quality images, we can conduct both Full-Reference (FR) and No-Reference (NR) evaluations. For FR metrics, we choose PSNR, SSIM, and LPIPS \cite{zhang2018unreasonable}. For NR metrics, we select MUSIQ \cite{ke2021musiq}, ClipIQA \cite{wang2023exploring}, and ManIQA \cite{yang2022maniqa}. RealPhoto60 \cite{yu2024scaling} is also used for testing with NR metrics, as the corresponding HQ images are not provided.

\begin{figure}[t]
    \centering
    \includegraphics[width=1.0\linewidth]{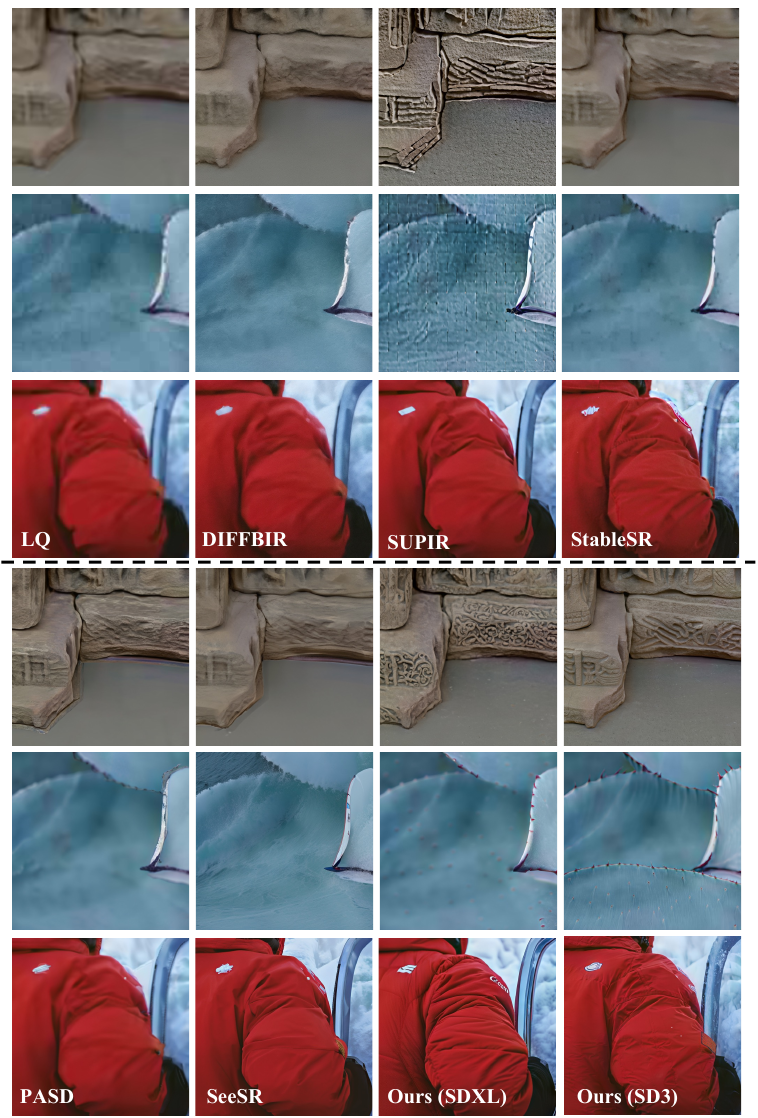}
    \caption{Qualitative comparisons on randomly cropped images from DIV2K validation set. From top to bottom, every three images represent the results of the corresponding methods.}
    \label{fig:rand_cmp}
\end{figure}

\begin{table*}[t]
\centering
\resizebox{0.9\linewidth}{!}{
    \begin{tabular}{c | c| c c c c c c c}
    \hline
    Datasets & \multicolumn{1}{c|}{Metrics} & StableSR & DiffBIR & PASD & SUPIR & SeeSR & Ours (SDXL) & Ours (SD3) \\
    \hline
    \multirow{6}{*}[-1.5ex]{\begin{tabular}[c]{@{}c@{}}DIV2K Valid\\ Center Cropped \\ (mixture degradation) \end{tabular}} 
    & ClipIQA &  0.4793  & 0.5633  & 0.6131 &  0.6504 & 0.6261    &  \underline{0.6517} & \textbf{0.6689 }\\
    & MUSIQ    & 63.08   & 68.79   & 70.27 &  70.02   & 71.84  & \underline{71.92}   & \textbf{72.78} \\
    & ManIQA   & 0.3255  & 0.3993  & 0.4632  & \underline{0.5182}  & 0.4734  &  0.4997  & \textbf{0.5217}  \\
    \cline{2-9}
    & PSNR     & \textbf{23.14} & \underline{22.84}  & 22.54    & 20.44     & 22.41  &  21.41    & 21.01 \\
    & SSIM     & \textbf{0.6526}& 0.598  & \underline{0.6152} & 0.5586     & 0.6086     & 0.576      & 0.584 \\
    & LPIPS $\downarrow$    & 0.3279    & 0.3163  & 0.3235     & 0.3307     & \textbf{0.3014} & 0.3187   &  \underline{0.3148} \\

    \hline
    \hline
    
    \multirow{6}{*}[-1.5ex]{\begin{tabular}[c]{@{}c@{}}DIV2K Valid\\ Center Cropped \\  (8$\times$ downsampling) \end{tabular}} 
    & ClipIQA & 0.563   & 0.6061  & 0.6116  & \underline{0.664}  & 0.6179 & 0.6449 & \textbf{0.6659} \\
    & MUSIQ    &  64.44   & 69.78  & 69.17  & 70.81  & 71.43 & \underline{71.92}  & \textbf{72.2}\\
    & ManIQA   & 0.3481   & 0.4191  & 0.4643  & \textbf{0.5311}  & 0.4652 &   0.4741     & \underline{0.5096} \\
    \cline{2-9}
    & PSNR     & \textbf{23.08}   & \underline{22.83}  & 22.52   & 20.04   & 22.28    & 21.35   & 20.68 \\
    & SSIM     & \textbf{0.644}   & 0.595  & \underline{0.6056}   & 0.5495   & 0.605 & 0.5765  & 0.5575 \\
    & LPIPS  $\downarrow$   &  0.3014  & \textbf{0.2841}  & 0.3057   & 0.3135  & \underline{0.3011}    & 0.3126        &  0.3224 \\

    \hline
    \hline
    
    \multirow{6}{*}[-1.5ex]{\begin{tabular}[c]{@{}c@{}}DIV2K Valid\\Randomly Cropped \\  (mixture degradation) \end{tabular}} 
    & ClipIQA & 0.3889   & 0.5530  & 0.5359  & 0.6356  & \underline{0.6804}  & 0.6703  & \textbf{0.6868}  \\
    & MUSIQ    &  43.25  & 57.41   &  59.46  & 68.25    & 69.42     & \underline{70.49}       & \textbf{70.56} \\
    & ManIQA   & 0.237   & 0.3709  & 0.3904   & 0.4740  & \underline{0.5106}  & \textbf{0.5358}   & 0.5047 \\
    \cline{2-9}
    & PSNR     & \textbf{23.29}  & \underline{23.02} & 22.85   & 20.83    & 22.01   & 21.31   & 21.14 \\
    & SSIM     & \textbf{0.6176}  & 0.5606  & \underline{0.5953} & 0.5309      & 0.5612    & 0.5412  & 0.5348 \\
    & LPIPS $\downarrow$ & 0.3896  & 0.3701  & 0.3983  & 0.4022    & \underline{0.3399 }& \textbf{0.334} &  0.3587 \\
    \hline

    \end{tabular}
}
\caption{Quantitative comparisons on the DIV2K Validation. For metrics marked with $\downarrow$, lower values indicate better performance, while for others, higher values are preferable.}
\label{tab:fr_cmp}
\end{table*}
\begin{table}[t]
\centering
\resizebox{1\linewidth}{!}{
    \begin{tabular}{c| c c c c}
    \hline
            Methods & \makecell{SDXL \\  ControlNet} & \makecell{SD3 \\  ControlNet} & Ours (SDXL) & Ours (SD3) \\
            \hline
            params & 839M & 504M & 157M & 80M \\
    \hline
    \end{tabular}
}
\caption{Comparison of parameter sizes between our methods and ControlNet.}
\label{tab:params}
\end{table}

\subsection{Compare with existing methods}
In this section, we compare our proposed methods with other recent state-of-the-art methods that also uses a diffusion prior. The methods includes SUPIR \cite{yu2024scaling}, StableSR \cite{wang2023exploiting}, PASD \cite{yang2023pixel}, SeeSR \cite{wu2024seesr} and DiffBIR \cite{lin2023diffbir}. Low quality input images are cropped to 1024 $\times$ 1024 for visual qualitative comparisons, the restoration results are resized to 512 $\times$ 512 for quantitative comparisons following \cite{yu2024scaling}, unless stated specifically. 

\subsubsection{Quantitative Comparisons}
We first present the quantitative comparisons on the RealPhoto60 and synthetic DIV2K validation sets. We used the PyIQA library \cite{pyiqa} to calculate all the metrics. Table \ref{tab:nr_cmp} shows a quantitative comparison of various methods on the RealPhoto60 dataset, measured using ClipIQA, MUSIQ, and ManIQA metrics. The results indicate that our proposed methods, ours (SDXL) and ours (SD3), perform competitively across all metrics. Specifically, our method (SD3) achieves the highest scores in ClipIQA (0.7067) and ManIQA (0.5292) , while ranking second in MUSIQ. Our method with SDXL achieves the highest score in MUSIQ(71.83). Overall, these results demonstrate the effectiveness of our approach in image restoration quality across different evaluation criteria. 

We now present the quantitative results for the DIV2K validation sets. The testing images are divided into center cropped (1024 $\times$ 1024) and randomly cropped (512 $\times$ 512) categories. For the center cropped images, we synthesize the LQ images using both mixture degradation and 8 $\times$ bilinear downsampling.
As shown in Table \ref{tab:fr_cmp}, for the center cropped part, our method on SD3 achieves the best results in all quality assessment metrics, and our method on SDXL secures second place in ClipIQA (0.6517) and MUSIQ (71.92). For the 8 $\times$ downsampled images, our method on SD3 achieves the best results in ClipIQA (0.6659) and MUSIQ (72.2), while our method on SDXL ranks second in MUSIQ (71.92). 
For the randomly cropped part, our method on SD3 achieves the best results in ClipIQA (0.6868) and MUSIQ (70.56). Our method on SDXL shows competitive performance, particularly in ManIQA (0.5358) and LPIPS (0.334), and ranks second in MUSIQ (70.49). Although our method does not outperform other methods in metrics like PSNR and SSIM, the visual quality results demonstrate that our approach maintains good visual fidelity. We will address this issue in the ablation study.

We also conduct a comparison on model size. Our models have trainable parameters of 157M (SDXL) and 80M (SD3), while ControlNet has approximately half the parameters of the original network. Methods like DiffBIR \cite{lin2023diffbir}, SeeSR \cite{wu2024seesr} and SUPIR \cite{yu2024scaling}, which utilize a ControlNet \cite{zhang2023adding} as LQ condition input, encounter problems when the prior's parameters increase, especially with priors like SD3 (2B), as shown in Table \ref{tab:params}. In contrast, our method uses only 80M parameters to achieve good restoration with SD3.

\subsubsection{Qualitative Comparisons}
For visual quality comparisons, we present the 1024 $\times$ 1024 results of different methods on the RealPhoto60 and DIV2K validation sets. The LQ images from the DIV2K validation set are generated through mixture degradations. As shown in Figure \ref{fig:main_cmp}, our methods present an excellent balance between fidelity and quality. 
SOTA methods like SUPIR and SeeSR can produce good restoration at most of the time, but sometimes they generate blurry results or results with less fidelity. For example, SUPIR produces incorrect textures for the stone (row 3) and blurry fur of a cat. SeeSR generates less detailed faces (row 1) and blurry results in rows 2 and 3. The last row shows that our methods can generate detailed textures for the hat and clothes, while other methods struggle to produce sufficient texture. Our results also show good visual fidelity, although we do not perform well on metrics like PSNR and SSIM, which suggests they are not robust metrics in terms of visual fidelity.

The last two columns display the results of our methods, showing sufficient details and sharp edges without obvious degradations, while maintaining visual fidelity. We also conducted experiments on close-up images created by randomly cropping the images from the DIV2K validation sets. In Figure \ref{fig:rand_cmp}, we observe that some methods produce either blurry results or incorrect textures, whereas our method consistently delivers high-quality images with rich details.

\begin{figure}[t]
    \centering
    \includegraphics[width=1.0\linewidth]{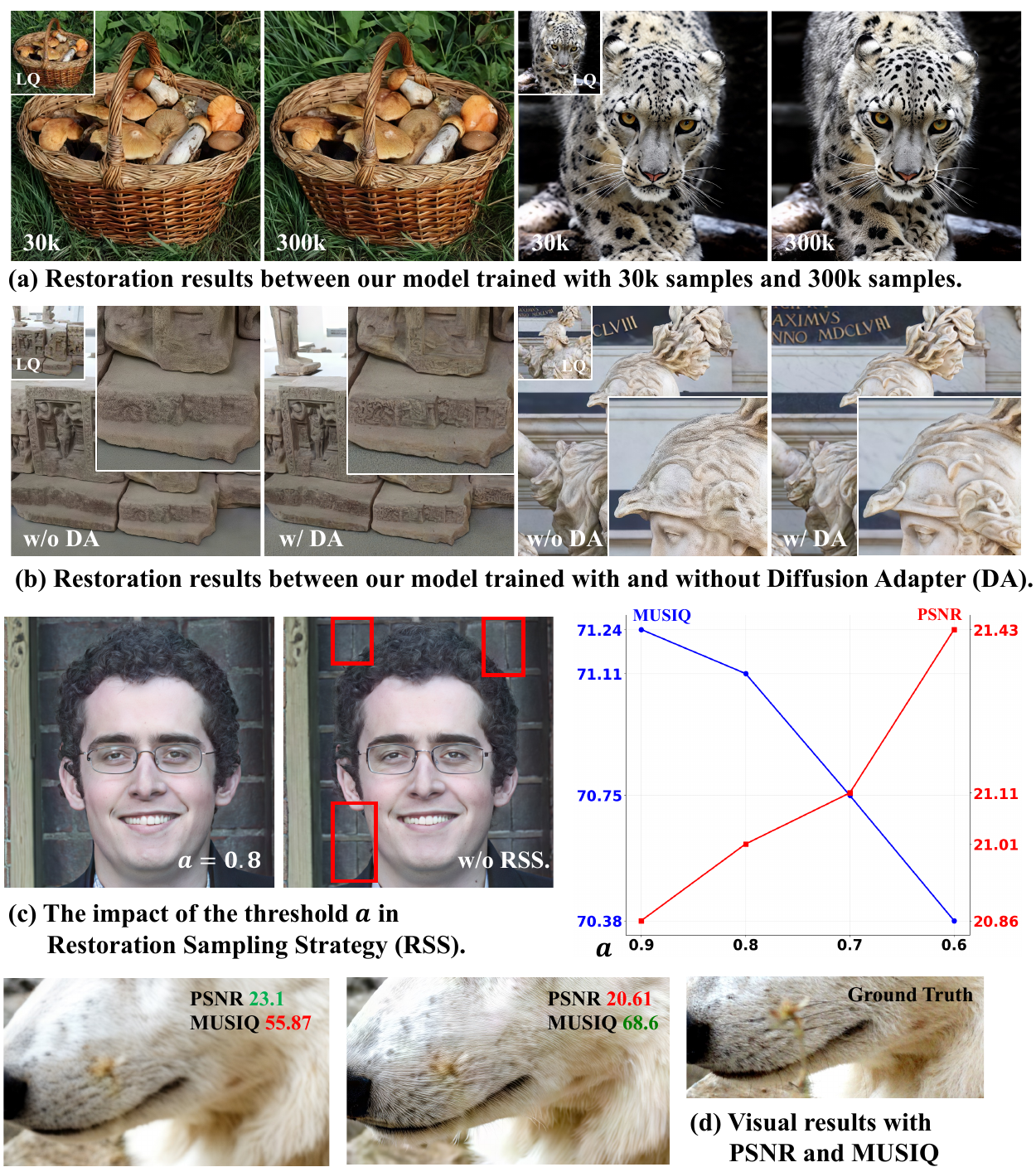}
    \caption{The figure illustrates the results of our model trained with 30k and 300k samples in (a). The qualitative comparisons on training with and without the Diffusion Adapter is shown in (b), and the impact of different values of threshold $a$ is shown in (c). }
    \label{fig:abl_n}
\end{figure}

\subsection{Ablation Study}
In this section, we present ablation studies only on our SDXL-based model, as our SD3-based model yields similar results.
\paragraph{Training Data Scale.} We trained our model using a dataset with 300k samples and a smaller one with 30k samples. The results are shown in Figure \ref {fig:abl_n} (a). While the overall quality is comparable, the model trained with 300k images exhibits higher quality and more detail. This demonstrates that scaling up the training data can yield better results, and our model can still achieve satisfactory performance even without a large-scale dataset.

\paragraph{Importance of the Diffusion Adapter.} We investigate the importance of the Diffusion Adapter in our model. We removed the Diffusion Adapter and trained the model under the same settings. As shown in Figure \ref {fig:abl_n} (b), the two samples demonstrate that without the Diffusion Adapter, the model can produce messy high-frequency details in some cases.

\paragraph{Restoration Sampling Strategy.}
Stochastic samplers can achieve high-quality generation. However, they may also introduce results that lack fidelity. We select stochastic sampler from \cite{karras2022elucidating} to investigate how threshold $a$ 
 of Restoration Sampling Strategy (RSS) balances the quality and fidelity. As shown in Figure \ref{fig:abl_n}, the result generated without RSS shows some unexpected details (circled in red). When the threshold $a$ is set to 0.8, the result appears more faithful to the LQ image (Please refer to row 1 in Figure \ref{fig:main_cmp}). Additionally, we evaluate the impact of different value of $a$ using 30 images from DIV2K mixture degradation set, and as shown in the right chart of Figure \ref{fig:abl_n} (c), quality scores delivered by MUSIQ increase as $a$ decreases. This is because the restoration can generate more high-frequency details. Some generated details might be unexpected, which is why we need RSS to impose constraints. Metrics like PSNR and SSIM aim to capture pixel-wise similarity, but they are not robust enough to capture visual fidelity to human eyes, as also reported in \cite{yu2024scaling, blau2018perception, gu2020pipal}. As shown in Figure \ref{fig:abl_n} (d), the blurred result can achieve a higher PSNR, while the result with more textures gets a lower PSNR. This is because the generated high-frequency textures from the diffusion model can not match the ground truth per pixel, even though these textures appear visually similar to the ground truth.

Additional visual results, including controllable restoration and a discussion on degradation-robust encoder in SDXL, are provided in supplementary material.

\section{Conclusion}
We designed a framework called the Diffusion Restoration Adapter, composed of Restoration Adapters and Diffusion Adapters, to utilize pretrained diffusion models as a restoration prior for real-world image restoration. This framework can work with either diffusion models that use a denoising UNet or DiTs. We specifically designed the architecture of the Restoration Adapter to accommodate both types of diffusion priors. Additionally, we propose the Restoration Sampling Strategy to balance quality and fidelity across different samplers. Quantitative and qualitative experiments have demonstrated the effectiveness of our method. 

{
    \small
    \bibliographystyle{ieeenat_fullname}
    \bibliography{main}

\begin{thebibliography}{60}
\providecommand{\natexlab}[1]{#1}
\providecommand{\url}[1]{\texttt{#1}}
\expandafter\ifx\csname urlstyle\endcsname\relax
  \providecommand{\doi}[1]{doi: #1}\else
  \providecommand{\doi}{doi: \begingroup \urlstyle{rm}\Url}\fi

\bibitem[Abdal et~al.(2019)Abdal, Qin, and Wonka]{abdal2019image2stylegan}
Rameen Abdal, Yipeng Qin, and Peter Wonka.
\newblock Image2stylegan: How to embed images into the stylegan latent space?
\newblock In \emph{Proceedings of the IEEE/CVF international conference on computer vision}, pages 4432--4441, 2019.

\bibitem[Abdal et~al.(2021)Abdal, Zhu, Mitra, and Wonka]{abdal2021styleflow}
Rameen Abdal, Peihao Zhu, Niloy~J Mitra, and Peter Wonka.
\newblock Styleflow: Attribute-conditioned exploration of stylegan-generated images using conditional continuous normalizing flows.
\newblock \emph{ACM Transactions on Graphics (ToG)}, 40\penalty0 (3):\penalty0 1--21, 2021.

\bibitem[Agustsson and Timofte(2017)]{Agustsson_2017_CVPR_Workshops}
Eirikur Agustsson and Radu Timofte.
\newblock Ntire 2017 challenge on single image super-resolution: Dataset and study.
\newblock In \emph{The IEEE Conference on Computer Vision and Pattern Recognition (CVPR) Workshops}, 2017.

\bibitem[Bau et~al.(2020)Bau, Strobelt, Peebles, Wulff, Zhou, Zhu, and Torralba]{bau2020semantic}
David Bau, Hendrik Strobelt, William Peebles, Jonas Wulff, Bolei Zhou, Jun-Yan Zhu, and Antonio Torralba.
\newblock Semantic photo manipulation with a generative image prior.
\newblock \emph{arXiv preprint arXiv:2005.07727}, 2020.

\bibitem[Blau and Michaeli(2018)]{blau2018perception}
Yochai Blau and Tomer Michaeli.
\newblock The perception-distortion tradeoff.
\newblock In \emph{Proceedings of the IEEE conference on computer vision and pattern recognition}, pages 6228--6237, 2018.

\bibitem[Chen and Mo(2022)]{pyiqa}
Chaofeng Chen and Jiadi Mo.
\newblock {IQA-PyTorch}: Pytorch toolbox for image quality assessment.
\newblock [Online]. Available: \url{https://github.com/chaofengc/IQA-PyTorch}, 2022.

\bibitem[Chen et~al.(2023)Chen, Yu, Ge, Yao, Xie, Wu, Wang, Kwok, Luo, Lu, et~al.]{chen2023pixart}
Junsong Chen, Jincheng Yu, Chongjian Ge, Lewei Yao, Enze Xie, Yue Wu, Zhongdao Wang, James Kwok, Ping Luo, Huchuan Lu, et~al.
\newblock Pixart-$\alpha$: Fast training of diffusion transformer for photorealistic text-to-image synthesis.
\newblock \emph{arXiv preprint arXiv:2310.00426}, 2023.

\bibitem[Chen et~al.(2024)Chen, Wu, Wang, Su, Chen, Xing, Zhong, Zhang, Zhu, Lu, et~al.]{chen2024internvl}
Zhe Chen, Jiannan Wu, Wenhai Wang, Weijie Su, Guo Chen, Sen Xing, Muyan Zhong, Qinglong Zhang, Xizhou Zhu, Lewei Lu, et~al.
\newblock Internvl: Scaling up vision foundation models and aligning for generic visual-linguistic tasks.
\newblock In \emph{Proceedings of the IEEE/CVF Conference on Computer Vision and Pattern Recognition}, pages 24185--24198, 2024.

\bibitem[Esser et~al.(2024)Esser, Kulal, Blattmann, Entezari, M{\"u}ller, Saini, Levi, Lorenz, Sauer, Boesel, et~al.]{esser2024scaling}
Patrick Esser, Sumith Kulal, Andreas Blattmann, Rahim Entezari, Jonas M{\"u}ller, Harry Saini, Yam Levi, Dominik Lorenz, Axel Sauer, Frederic Boesel, et~al.
\newblock Scaling rectified flow transformers for high-resolution image synthesis.
\newblock In \emph{Forty-first International Conference on Machine Learning}, 2024.

\bibitem[Goodfellow et~al.(2014)Goodfellow, Pouget-Abadie, Mirza, Xu, Warde-Farley, Ozair, Courville, and Bengio]{goodfellow2014generative}
Ian Goodfellow, Jean Pouget-Abadie, Mehdi Mirza, Bing Xu, David Warde-Farley, Sherjil Ozair, Aaron Courville, and Yoshua Bengio.
\newblock Generative adversarial nets.
\newblock \emph{Advances in neural information processing systems}, 27, 2014.

\bibitem[Gu et~al.(2020{\natexlab{a}})Gu, Cai, Chen, Ye, Ren, and Dong]{gu2020pipal}
Jinjin Gu, Haoming Cai, Haoyu Chen, Xiaoxing Ye, Jimmy Ren, and Chao Dong.
\newblock Pipal: a large-scale image quality assessment dataset for perceptual image restoration.
\newblock In \emph{Computer Vision--ECCV 2020: 16th European Conference, Glasgow, UK, August 23--28, 2020, Proceedings, Part XI 16}, pages 633--651, 2020{\natexlab{a}}.

\bibitem[Gu et~al.(2020{\natexlab{b}})Gu, Shen, and Zhou]{gu2020image}
Jinjin Gu, Yujun Shen, and Bolei Zhou.
\newblock Image processing using multi-code gan prior.
\newblock In \emph{Proceedings of the IEEE/CVF conference on computer vision and pattern recognition}, pages 3012--3021, 2020{\natexlab{b}}.

\bibitem[H{\"a}rk{\"o}nen et~al.(2020)H{\"a}rk{\"o}nen, Hertzmann, Lehtinen, and Paris]{harkonen2020ganspace}
Erik H{\"a}rk{\"o}nen, Aaron Hertzmann, Jaakko Lehtinen, and Sylvain Paris.
\newblock Ganspace: Discovering interpretable gan controls.
\newblock \emph{Advances in neural information processing systems}, 33:\penalty0 9841--9850, 2020.

\bibitem[Ho and Salimans(2022)]{ho2022classifier}
Jonathan Ho and Tim Salimans.
\newblock Classifier-free diffusion guidance.
\newblock \emph{arXiv preprint arXiv:2207.12598}, 2022.

\bibitem[Ho et~al.(2020)Ho, Jain, and Abbeel]{ho2020denoising}
Jonathan Ho, Ajay Jain, and Pieter Abbeel.
\newblock Denoising diffusion probabilistic models.
\newblock \emph{Advances in neural information processing systems}, 33:\penalty0 6840--6851, 2020.

\bibitem[Hu et~al.(2021)Hu, Shen, Wallis, Allen-Zhu, Li, Wang, Wang, and Chen]{hu2021lora}
Edward~J Hu, Yelong Shen, Phillip Wallis, Zeyuan Allen-Zhu, Yuanzhi Li, Shean Wang, Lu Wang, and Weizhu Chen.
\newblock Lora: Low-rank adaptation of large language models.
\newblock \emph{arXiv preprint arXiv:2106.09685}, 2021.

\bibitem[Karras et~al.(2019)Karras, Laine, and Aila]{karras2019style}
Tero Karras, Samuli Laine, and Timo Aila.
\newblock A style-based generator architecture for generative adversarial networks.
\newblock In \emph{Proceedings of the IEEE/CVF conference on computer vision and pattern recognition}, pages 4401--4410, 2019.

\bibitem[Karras et~al.(2020)Karras, Laine, Aittala, Hellsten, Lehtinen, and Aila]{karras2020analyzing}
Tero Karras, Samuli Laine, Miika Aittala, Janne Hellsten, Jaakko Lehtinen, and Timo Aila.
\newblock Analyzing and improving the image quality of stylegan.
\newblock In \emph{Proceedings of the IEEE/CVF conference on computer vision and pattern recognition}, pages 8110--8119, 2020.

\bibitem[Karras et~al.(2022)Karras, Aittala, Aila, and Laine]{karras2022elucidating}
Tero Karras, Miika Aittala, Timo Aila, and Samuli Laine.
\newblock Elucidating the design space of diffusion-based generative models.
\newblock \emph{Advances in Neural Information Processing Systems}, 35:\penalty0 26565--26577, 2022.

\bibitem[Kawar et~al.(2022)Kawar, Elad, Ermon, and Song]{kawar2022denoising}
Bahjat Kawar, Michael Elad, Stefano Ermon, and Jiaming Song.
\newblock Denoising diffusion restoration models.
\newblock \emph{Advances in Neural Information Processing Systems}, 35:\penalty0 23593--23606, 2022.

\bibitem[Ke et~al.(2021)Ke, Wang, Wang, Milanfar, and Yang]{ke2021musiq}
Junjie Ke, Qifei Wang, Yilin Wang, Peyman Milanfar, and Feng Yang.
\newblock Musiq: Multi-scale image quality transformer.
\newblock In \emph{Proceedings of the IEEE/CVF International Conference on Computer Vision}, pages 5148--5157, 2021.

\bibitem[Kingma(2013)]{kingma2013auto}
Diederik~P Kingma.
\newblock Auto-encoding variational bayes.
\newblock \emph{arXiv preprint arXiv:1312.6114}, 2013.

\bibitem[Li et~al.(2024)Li, Zhang, Lin, Xiong, Long, Deng, Zhang, Liu, Huang, Xiao, et~al.]{li2024hunyuan}
Zhimin Li, Jianwei Zhang, Qin Lin, Jiangfeng Xiong, Yanxin Long, Xinchi Deng, Yingfang Zhang, Xingchao Liu, Minbin Huang, Zedong Xiao, et~al.
\newblock Hunyuan-dit: A powerful multi-resolution diffusion transformer with fine-grained chinese understanding.
\newblock \emph{arXiv preprint arXiv:2405.08748}, 2024.

\bibitem[Liang et~al.(2021{\natexlab{a}})Liang, Hou, and Shen]{liang2021ssflow}
Hanbang Liang, Xianxu Hou, and Linlin Shen.
\newblock Ssflow: style-guided neural spline flows for face image manipulation.
\newblock In \emph{Proceedings of the 29th ACM International Conference on Multimedia}, pages 79--87, 2021{\natexlab{a}}.

\bibitem[Liang et~al.(2021{\natexlab{b}})Liang, Cao, Sun, Zhang, Van~Gool, and Timofte]{liang2021swinir}
Jingyun Liang, Jiezhang Cao, Guolei Sun, Kai Zhang, Luc Van~Gool, and Radu Timofte.
\newblock Swinir: Image restoration using swin transformer.
\newblock In \emph{Proceedings of the IEEE/CVF international conference on computer vision}, pages 1833--1844, 2021{\natexlab{b}}.

\bibitem[Lin et~al.(2023)Lin, He, Chen, Lyu, Fei, Dai, Ouyang, Qiao, and Dong]{lin2023diffbir}
Xinqi Lin, Jingwen He, Ziyan Chen, Zhaoyang Lyu, Ben Fei, Bo Dai, Wanli Ouyang, Yu Qiao, and Chao Dong.
\newblock Diffbir: Towards blind image restoration with generative diffusion prior.
\newblock \emph{arXiv preprint arXiv:2308.15070}, 2023.

\bibitem[Lipman et~al.(2023)Lipman, Chen, Ben-Hamu, Nickel, and Le]{lipmanflow}
Yaron Lipman, Ricky~TQ Chen, Heli Ben-Hamu, Maximilian Nickel, and Matthew Le.
\newblock Flow matching for generative modeling.
\newblock In \emph{The Eleventh International Conference on Learning Representations (ICLR)}, 2023.

\bibitem[Loshchilov and Hutter(2017)]{loshchilov2017decoupled}
Ilya Loshchilov and Frank Hutter.
\newblock Decoupled weight decay regularization.
\newblock \emph{arXiv preprint arXiv:1711.05101}, 2017.

\bibitem[Lu et~al.(2022{\natexlab{a}})Lu, Zhou, Bao, Chen, Li, and Zhu]{lu2022dpm}
Cheng Lu, Yuhao Zhou, Fan Bao, Jianfei Chen, Chongxuan Li, and Jun Zhu.
\newblock Dpm-solver: A fast ode solver for diffusion probabilistic model sampling in around 10 steps.
\newblock \emph{Advances in Neural Information Processing Systems}, 35:\penalty0 5775--5787, 2022{\natexlab{a}}.

\bibitem[Lu et~al.(2022{\natexlab{b}})Lu, Zhou, Bao, Chen, Li, and Zhu]{lu2022dpmpp}
Cheng Lu, Yuhao Zhou, Fan Bao, Jianfei Chen, Chongxuan Li, and Jun Zhu.
\newblock Dpm-solver++: Fast solver for guided sampling of diffusion probabilistic models.
\newblock \emph{arXiv preprint arXiv:2211.01095}, 2022{\natexlab{b}}.

\bibitem[Menon et~al.(2020)Menon, Damian, Hu, Ravi, and Rudin]{menon2020pulse}
Sachit Menon, Alexandru Damian, Shijia Hu, Nikhil Ravi, and Cynthia Rudin.
\newblock Pulse: Self-supervised photo upsampling via latent space exploration of generative models.
\newblock In \emph{Proceedings of the ieee/cvf conference on computer vision and pattern recognition}, pages 2437--2445, 2020.

\bibitem[Pan et~al.(2021)Pan, Zhan, Dai, Lin, Loy, and Luo]{pan2021exploiting}
Xingang Pan, Xiaohang Zhan, Bo Dai, Dahua Lin, Chen~Change Loy, and Ping Luo.
\newblock Exploiting deep generative prior for versatile image restoration and manipulation.
\newblock \emph{IEEE Transactions on Pattern Analysis and Machine Intelligence}, 44\penalty0 (11):\penalty0 7474--7489, 2021.

\bibitem[Peebles and Xie(2023)]{peebles2023scalable}
William Peebles and Saining Xie.
\newblock Scalable diffusion models with transformers.
\newblock In \emph{Proceedings of the IEEE/CVF International Conference on Computer Vision}, pages 4195--4205, 2023.

\bibitem[Podell et~al.(2023)Podell, English, Lacey, Blattmann, Dockhorn, M{\"u}ller, Penna, and Rombach]{podell2023sdxl}
Dustin Podell, Zion English, Kyle Lacey, Andreas Blattmann, Tim Dockhorn, Jonas M{\"u}ller, Joe Penna, and Robin Rombach.
\newblock Sdxl: Improving latent diffusion models for high-resolution image synthesis.
\newblock \emph{arXiv preprint arXiv:2307.01952}, 2023.

\bibitem[Rombach et~al.(2022)Rombach, Blattmann, Lorenz, Esser, and Ommer]{rombach2022high}
Robin Rombach, Andreas Blattmann, Dominik Lorenz, Patrick Esser, and Bj{\"o}rn Ommer.
\newblock High-resolution image synthesis with latent diffusion models.
\newblock In \emph{Proceedings of the IEEE/CVF conference on computer vision and pattern recognition}, pages 10684--10695, 2022.

\bibitem[Schuhmann et~al.(2022)Schuhmann, Beaumont, Vencu, Gordon, Wightman, Cherti, Coombes, Katta, Mullis, Wortsman, et~al.]{schuhmann2022laion}
Christoph Schuhmann, Romain Beaumont, Richard Vencu, Cade Gordon, Ross Wightman, Mehdi Cherti, Theo Coombes, Aarush Katta, Clayton Mullis, Mitchell Wortsman, et~al.
\newblock Laion-5b: An open large-scale dataset for training next generation image-text models.
\newblock \emph{Advances in Neural Information Processing Systems}, 35:\penalty0 25278--25294, 2022.

\bibitem[Shen et~al.(2020)Shen, Yang, Tang, and Zhou]{shen2020interfacegan}
Yujun Shen, Ceyuan Yang, Xiaoou Tang, and Bolei Zhou.
\newblock Interfacegan: Interpreting the disentangled face representation learned by gans.
\newblock \emph{IEEE transactions on pattern analysis and machine intelligence}, 44\penalty0 (4):\penalty0 2004--2018, 2020.

\bibitem[Song et~al.(2021)Song, Meng, and Ermon]{songdenoising}
Jiaming Song, Chenlin Meng, and Stefano Ermon.
\newblock Denoising diffusion implicit models.
\newblock In \emph{International Conference on Learning Representations (ICLR)}, 2021.

\bibitem[Song and Ermon(2019)]{song2019generative}
Yang Song and Stefano Ermon.
\newblock Generative modeling by estimating gradients of the data distribution.
\newblock \emph{Advances in neural information processing systems}, 32, 2019.

\bibitem[Song et~al.(2020)Song, Sohl-Dickstein, Kingma, Kumar, Ermon, and Poole]{song2020score}
Yang Song, Jascha Sohl-Dickstein, Diederik~P Kingma, Abhishek Kumar, Stefano Ermon, and Ben Poole.
\newblock Score-based generative modeling through stochastic differential equations.
\newblock \emph{arXiv preprint arXiv:2011.13456}, 2020.

\bibitem[Tewari et~al.(2020)Tewari, Elgharib, Bharaj, Bernard, Seidel, P{\'e}rez, Zollhofer, and Theobalt]{tewari2020stylerig}
Ayush Tewari, Mohamed Elgharib, Gaurav Bharaj, Florian Bernard, Hans-Peter Seidel, Patrick P{\'e}rez, Michael Zollhofer, and Christian Theobalt.
\newblock Stylerig: Rigging stylegan for 3d control over portrait images.
\newblock In \emph{Proceedings of the IEEE/CVF conference on computer vision and pattern recognition}, pages 6142--6151, 2020.

\bibitem[Wang et~al.(2023{\natexlab{a}})Wang, Chan, and Loy]{wang2023exploring}
Jianyi Wang, Kelvin~CK Chan, and Chen~Change Loy.
\newblock Exploring clip for assessing the look and feel of images.
\newblock In \emph{Proceedings of the AAAI Conference on Artificial Intelligence}, pages 2555--2563, 2023{\natexlab{a}}.

\bibitem[Wang et~al.(2023{\natexlab{b}})Wang, Yue, Zhou, Chan, and Loy]{wang2023exploiting}
Jianyi Wang, Zongsheng Yue, Shangchen Zhou, Kelvin~CK Chan, and Chen~Change Loy.
\newblock Exploiting diffusion prior for real-world image super-resolution.
\newblock \emph{arXiv preprint arXiv:2305.07015}, 2023{\natexlab{b}}.

\bibitem[Wang et~al.(2021{\natexlab{a}})Wang, Li, Zhang, and Shan]{wang2021towards}
Xintao Wang, Yu Li, Honglun Zhang, and Ying Shan.
\newblock Towards real-world blind face restoration with generative facial prior.
\newblock In \emph{Proceedings of the IEEE/CVF conference on computer vision and pattern recognition}, pages 9168--9178, 2021{\natexlab{a}}.

\bibitem[Wang et~al.(2021{\natexlab{b}})Wang, Xie, Dong, and Shan]{wang2021real}
Xintao Wang, Liangbin Xie, Chao Dong, and Ying Shan.
\newblock Real-esrgan: Training real-world blind super-resolution with pure synthetic data.
\newblock In \emph{Proceedings of the IEEE/CVF International Conference on Computer Vision}, pages 1905--1914, 2021{\natexlab{b}}.

\bibitem[Wu et~al.(2024)Wu, Yang, Sun, Zhang, Li, and Zhang]{wu2024seesr}
Rongyuan Wu, Tao Yang, Lingchen Sun, Zhengqiang Zhang, Shuai Li, and Lei Zhang.
\newblock Seesr: Towards semantics-aware real-world image super-resolution.
\newblock In \emph{Proceedings of the IEEE/CVF conference on computer vision and pattern recognition}, pages 25456--25467, 2024.

\bibitem[Yang et~al.(2022)Yang, Wu, Shi, Lao, Gong, Cao, Wang, and Yang]{yang2022maniqa}
Sidi Yang, Tianhe Wu, Shuwei Shi, Shanshan Lao, Yuan Gong, Mingdeng Cao, Jiahao Wang, and Yujiu Yang.
\newblock Maniqa: Multi-dimension attention network for no-reference image quality assessment.
\newblock In \emph{Proceedings of the IEEE/CVF Conference on Computer Vision and Pattern Recognition}, pages 1191--1200, 2022.

\bibitem[Yang et~al.(2021)Yang, Ren, Xie, and Zhang]{yang2021gan}
Tao Yang, Peiran Ren, Xuansong Xie, and Lei Zhang.
\newblock Gan prior embedded network for blind face restoration in the wild.
\newblock In \emph{Proceedings of the IEEE/CVF Conference on Computer Vision and Pattern Recognition}, pages 672--681, 2021.

\bibitem[Yang et~al.(2023)Yang, Ren, Xie, and Zhang]{yang2023pixel}
Tao Yang, Peiran Ren, Xuansong Xie, and Lei Zhang.
\newblock Pixel-aware stable diffusion for realistic image super-resolution and personalized stylization.
\newblock \emph{arXiv preprint arXiv:2308.14469}, 2023.

\bibitem[Yu et~al.(2024)Yu, Gu, Li, Hu, Kong, Wang, He, Qiao, and Dong]{yu2024scaling}
Fanghua Yu, Jinjin Gu, Zheyuan Li, Jinfan Hu, Xiangtao Kong, Xintao Wang, Jingwen He, Yu Qiao, and Chao Dong.
\newblock Scaling up to excellence: Practicing model scaling for photo-realistic image restoration in the wild.
\newblock In \emph{Proceedings of the IEEE/CVF Conference on Computer Vision and Pattern Recognition}, pages 25669--25680, 2024.

\bibitem[Zamir et~al.(2022)Zamir, Arora, Khan, Hayat, Khan, and Yang]{Zamir_2022_CVPR}
Syed~Waqas Zamir, Aditya Arora, Salman Khan, Munawar Hayat, Fahad~Shahbaz Khan, and Ming-Hsuan Yang.
\newblock Restormer: Efficient transformer for high-resolution image restoration.
\newblock In \emph{Proceedings of the IEEE/CVF Conference on Computer Vision and Pattern Recognition (CVPR)}, pages 5728--5739, 2022.

\bibitem[Zhang et~al.(2017)Zhang, Zuo, Gu, and Zhang]{zhang2017learning}
Kai Zhang, Wangmeng Zuo, Shuhang Gu, and Lei Zhang.
\newblock Learning deep cnn denoiser prior for image restoration.
\newblock In \emph{Proceedings of the IEEE conference on computer vision and pattern recognition}, pages 3929--3938, 2017.

\bibitem[Zhang et~al.(2021{\natexlab{a}})Zhang, Li, Zuo, Zhang, Van~Gool, and Timofte]{zhang2021plug}
Kai Zhang, Yawei Li, Wangmeng Zuo, Lei Zhang, Luc Van~Gool, and Radu Timofte.
\newblock Plug-and-play image restoration with deep denoiser prior.
\newblock \emph{IEEE Transactions on Pattern Analysis and Machine Intelligence}, 44\penalty0 (10):\penalty0 6360--6376, 2021{\natexlab{a}}.

\bibitem[Zhang et~al.(2021{\natexlab{b}})Zhang, Liang, Van~Gool, and Timofte]{zhang2021blind}
Kai Zhang, Jingyun Liang, Luc Van~Gool, and Radu Timofte.
\newblock Designing a practical degradation model for deep blind image super-resolution.
\newblock In \emph{Proceedings of the IEEE/CVF International Conference on Computer Vision}, pages 4791--4800, 2021{\natexlab{b}}.

\bibitem[Zhang et~al.(2023)Zhang, Rao, and Agrawala]{zhang2023adding}
Lvmin Zhang, Anyi Rao, and Maneesh Agrawala.
\newblock Adding conditional control to text-to-image diffusion models.
\newblock In \emph{Proceedings of the IEEE/CVF International Conference on Computer Vision}, pages 3836--3847, 2023.

\bibitem[Zhang et~al.(2018)Zhang, Isola, Efros, Shechtman, and Wang]{zhang2018unreasonable}
Richard Zhang, Phillip Isola, Alexei~A Efros, Eli Shechtman, and Oliver Wang.
\newblock The unreasonable effectiveness of deep features as a perceptual metric.
\newblock In \emph{Proceedings of the IEEE conference on computer vision and pattern recognition}, pages 586--595, 2018.

\bibitem[Zhao et~al.(2024)Zhao, Bai, Rao, Zhou, and Lu]{zhao2024unipc}
Wenliang Zhao, Lujia Bai, Yongming Rao, Jie Zhou, and Jiwen Lu.
\newblock Unipc: A unified predictor-corrector framework for fast sampling of diffusion models.
\newblock \emph{Advances in Neural Information Processing Systems}, 36, 2024.

\bibitem[Zhao et~al.(2022)Zhao, Su, Chu, Li, Renn, Zhu, Chen, and Jia]{zhao2022rethinking}
Yang Zhao, Yu-Chuan Su, Chun-Te Chu, Yandong Li, Marius Renn, Yukun Zhu, Changyou Chen, and Xuhui Jia.
\newblock Rethinking deep face restoration.
\newblock In \emph{Proceedings of the IEEE/CVF Conference on Computer Vision and Pattern Recognition}, pages 7652--7661, 2022.

\bibitem[Zhou et~al.(2022)Zhou, Chan, Li, and Loy]{zhou2022towards}
Shangchen Zhou, Kelvin Chan, Chongyi Li, and Chen~Change Loy.
\newblock Towards robust blind face restoration with codebook lookup transformer.
\newblock \emph{Advances in Neural Information Processing Systems}, 35:\penalty0 30599--30611, 2022.

\bibitem[Zhou et~al.(2024)Zhou, Chen, Pan, Shi, and Yang]{Zhou_2024_CVPR}
Shihao Zhou, Duosheng Chen, Jinshan Pan, Jinglei Shi, and Jufeng Yang.
\newblock Adapt or perish: Adaptive sparse transformer with attentive feature refinement for image restoration.
\newblock In \emph{Proceedings of the IEEE/CVF Conference on Computer Vision and Pattern Recognition (CVPR)}, pages 2952--2963, 2024.

\end{thebibliography}
}


\end{document}